\documentclass[journal]{IEEEtran}

\usepackage{ulem}
\usepackage{times}
\usepackage{epsfig}
\usepackage{amsmath}
\usepackage{amssymb}
\usepackage{amsmath}

\usepackage[pagebackref=true,breaklinks=true,letterpaper=true,colorlinks,bookmarks=false]{hyperref}

\usepackage{enumerate}
\usepackage{multirow}  
\usepackage{stfloats}
\usepackage{booktabs}
\usepackage{diagbox}
\usepackage{makecell}

\usepackage{graphics}
\usepackage{subfigure}
\graphicspath{{figs/}}

\usepackage{algorithm}
\usepackage{algorithmic}
\usepackage{color}
\usepackage{colortbl}
\usepackage{comment}
\definecolor{LightRed}{rgb}{1,0.92,0.92}
\definecolor{LightOrange}{rgb}{1,0.95,0.88}
\definecolor{LightYellow}{rgb}{1.0,1.0,0.84}
\definecolor{LightGreen}{rgb}{0.9,1.0,0.88}
\definecolor{LightCyan}{rgb}{0.9,1,1}
\definecolor{LightBlue}{rgb}{0.9,0.94,1}
\definecolor{LightIndigo}{rgb}{0.92,0.9,1}
\definecolor{LightMagenta}{rgb}{0.96,0.86,1}
\definecolor{DirtyWhite}{rgb}{0.96,0.96,0.96}

\usepackage[x11names]{xcolor}
\usepackage[
     colorlinks,
]{hyperref}

\usepackage{makecell}

\begin{document}

\title{RSGPT: A Remote Sensing Vision Language Model and Benchmark}

\author{Yuan Hu$^\dagger$, Jianlong Yuan$^\dagger$, Congcong Wen, \textit{IEEE Member}, Xiaonan Lu, Xiang Li$^*$, \textit{IEEE Member}.
\thanks{Yuan Hu and Jianlong Yuan are with the DAMO Academy, Alibaba Group, Beijing, China (email: huyuan826@gmail.com, gongyuan.yjl@alibaba-inc.com). Congcong Wen is with the Department of Electrical and Computer Engineering, New York University Abu Dhabi, Abu Dhabi, UAE (email: wencc@nyu.edu). Xiaonan Lu is with the University of Chinese Academy of Sciences, Beijing, China (email: email: luxiaonan96@gmail.com). Xiang Li is with the King Abdullah University of Science and Technology, Jeddah, Saudi Arabia (email: xiangli92@ieee.org). 
}
\thanks{$\dagger$ The authors contributed equally to this work.}
\thanks{$*$ Corresponding author: Xiang Li.}
}


\maketitle

\begin{abstract}
The emergence of large-scale Large Language Models (LLMs), with GPT-4 as a prominent example, has significantly propelled the rapid advancement of Artificial General Intelligence (AGI) and sparked the revolution of Artificial Intelligence 2.0. 
In the realm of remote sensing, there is a growing interest in developing large vision language models (VLMs) specifically tailored for data analysis in this domain. However, current research predominantly revolves around visual recognition tasks, lacking comprehensive, large-scale image-text datasets that are aligned and suitable for training large VLMs, which poses significant challenges to effectively training such models for remote sensing applications.
In computer vision, recent research has demonstrated that fine-tuning large vision language models on small-scale, high-quality datasets can yield impressive performance in visual and language understanding. These results are comparable to state-of-the-art VLMs trained from scratch on massive amounts of data, such as GPT-4.
Inspired by this captivating idea, in this work, we build a high-quality Remote Sensing Image Captioning dataset (\textbf{RSICap}) that facilitates the development of large VLMs in the remote sensing field. Unlike previous remote sensing datasets that either employ model-generated captions or short descriptions, RSICap comprises 2,585 human-annotated captions with rich and high-quality information. This dataset offers detailed descriptions for each image, encompassing scene descriptions (e.g., residential area, airport, or farmland) as well as object information (e.g., color, shape, quantity, absolute position, etc). To facilitate the evaluation of VLMs in the field of remote sensing, we also provide a benchmark evaluation dataset called \textbf{RSIEval}. This dataset consists of human-annotated captions and visual question-answer pairs, allowing for a comprehensive assessment of VLMs in the context of remote sensing. We are actively engaged in expanding the scale of these two datasets to cover a broader spectrum of remote sensing image understanding tasks, further enhancing their utility and applicability.
Our dataset and codes will be released at \url{https://github.com/Lavender105/RSGPT}.
\end{abstract}

\begin{IEEEkeywords}
Remote Sensing, Genrative Pretrained Transformer (GPT), Large Language Model (LLM), Vision Language Model (VLM)
\end{IEEEkeywords}

\section{Introduction}
Deep learning models, such as convolutional neural networks (CNNs)~\cite{lenet} and Vision Transformers~\cite{vit}, are capable of automatically learning complex patterns and representations from remote sensing data. It enables accurate and efficient analysis of large-scale remote sensing datasets and now has become the dominant method for remote sensing data analysis. The use of deep learning in remote sensing has significantly contributed to improved understanding and decision-making in various geo-related applications, including environmental monitoring, agriculture, urban planning, and disaster management~\cite{li2018building,wen2019novel,li2021geometry,tran2020damage}. A comprehensive review of deep learning in remote sensing can be found at~\cite{zhu2017deep,zhang2016deep,ma2019deep}. Although deep learning methods have achieved success in various remote sensing image analysis tasks, existing deep learning-based models mostly focus on the visual information extraction from remote sensing images, while paying less attention to the semantic understanding of visual tokens and lack of general reasoning abilities~\cite{wen2023vision}.

In the past few years, Large language models (LLMs), built from large transformer networks have achieved remarkable performance in natural language understanding and generation tasks, such as language modeling, text generation, and question answering~\cite{radford2018improving}. Of particular note are the advancements showcased by ChatGPT~\cite{chatgpt} and GPT-4~\cite{openai2023gpt4}. These models have showcased astonishing levels of general world knowledge and reasoning capabilities, achieved through extensive learning from vast amounts of textual data. Their achievements in language understanding have significantly contributed to the emergence of what is known as AI2.0. Recent research has witnessed a growing interest in integrating visual models, such as ResNet~\cite{resnet} and ViT~\cite{vit}, with LLMs to create a new class of models referred to as Vision Language Models (VLMs). 

VLMs refer to a category of artificial intelligence models that integrate computer vision and natural language processing techniques to achieve a holistic comprehension of visual and textual data. By concurrently analyzing visual and semantic patterns, VLMs possess the capability to discern intricate relationships between visual elements and linguistic information and offer a more comprehensive and human-like ability to understand visual content. In the past few years, researchers have explored VLMs for numerous remote sensing tasks, including image captioning~\cite{shi2017can,lu2017exploring,zhang2017natural,zhang2019description,li2020multi,wang2020word,li2020truncation,zhao2021high,zia2022transforming}, text-based image generation~\cite{bejiga2019retro,chen2021remote,zhao2021text,hochreiter1997long,xu2022txt2img}, text-based image retrieval~\cite{abdullah2020textrs,hochreiter1997long,rahhal2020deep,hochreiter1997long,yuan2022remote,al2022multilanguage,cheng2021deep,yuan2021lightweight,yuan2022exploring,rahhal2023contrasting}, visual question answering~\cite{lobry2020rsvqa,zheng2021mutual,chappuis2022language,chappuis2022language,al2022open,bazi2022bi,yuan2022easy,yuan2022change}, scene classification~\cite{li2017zero,sumbul2017fine,quan2018structural,wang2021distance,li2022generative}, etc. 
A comprehensive review can be found at~\cite{wen2023vision}. 

While these early attempts have shown the success of applying VLMs to remote sensing, it is still an emerging field with many unsolved challenges. Existing VLMs in the remote sensing field mostly still focus on developing more powerful visual foundation models to better extract more representative and robust visual features from remote sensing images (RSIs), thus lacking general reasoning abilities that are emergent in LLMs. Another challenging issue that impedes the development of VLMs in remote sensing is the lack of large-scale aligned image-text datasets. Existing RSI datasets mostly focus on visual recognition tasks and do not provide language annotations. Only a few attempts~\cite{ucm_sydney_caption,ucm_sydney_caption,nwpu_caption,zhang2023rs5m} tried to build image-text RSI datasets but data scale and quality are far from expected.

Recent work~\cite{zhou2023lima} indicates that when training LLMs, not only data quantity but rather prompt diversity and response quality really matter for building a high-performance LLM. Their experiments show that finetuning a 65B parameter LLaMA on a small-scale high-quality dataset with only 1,000 prompt-response can lead to a remarkably strong performance, comparable with GPT-4. 

Inspired by this exciting finding~\cite{zhou2023lima}, we build a high-quality Remote Sensing Image Captioning dataset (\textbf{RSICap}) to facilitate the construction of large vision language models in the remote sensing field. Note that concurrent work~\cite{zhang2023rs5m} builds a large-scale remote sensing vision-language dataset with 5 million aligned image-text pairs from model-generated captions. In contrast, our RSICap dataset collects 2,585 human-annotated captions with high-quality and rich information. Instead of using short descriptions like the previous dataset, our RSICap dataset provides human-annotated detailed descriptions for each image, covering scene descriptions (e.g., residential area, airport, or farmland), object information (color, shape, counting, absolute position), object relationship (e.g., relative position), and also visual reasoning knowledge (e.g., image capture season). This high-quality dataset facilitates finetuning existing large VLMs to build domain-specific VLMs in remote sensing. We also provide an evaluation dataset (\textbf{RSIEval}) dataset that can be used for the evaluation of domain-specific or general VLMs. RSIEval consists of 100 human-annotated captions and 936 visual question-answer pairs with rich information and open-ended questions and answers. We target building a standard benchmark that covers diverse remote sensing image understanding tasks, including image captioning, visual question answering, visual grounding, etc. In the current version, we benchmark two tasks, i.e., image captioning and visual question answering, using cutting-edge VLMs on our RSIEval dataset.

Recent work MiniGPT-4~\cite{zhu2023minigpt} shows the possibility of training 
a single projection layer can well align visual features with LLMs and the model shows powerful visual and language understanding capabilities like GPT-4 and also new emerging capabilities including writing stories and poems. Inspired by the great success of MiniGPT-4~\cite{zhu2023minigpt}, we develop a Remote Sensing Generative Pretrained Model (\textbf{RSGPT}) based on finetuning InstructBLIP~\cite{dai2023instructblip} on our newly created RSICap dataset. By finetuning only the Q-Former network and the linear layer of InstructBLIP, our model can quickly learn to align visual features of remote sensing images with LLMs in a data-efficient way. Our RSGPT model shows better image captioning and visual question-answering performance than previous SOTA methods.

\section{Related Work}
\subsection{GPT}
Generative Pre-trained Transformer (GPT) is currently the most popular and successful model in natural language processing and computer vision. The first version of GPT~\cite{radford2018improving} model developed by OpenAI, is one of the pioneering works that explore large transformer networks for diverse natural language processing tasks. By training on web-scale text datasets, this model achieved impressive performance on a wide range of language modeling and text generation tasks. Following the great success of the GPT model, GPT-2 further increased the model capacity to 1.5 billion parameters and demonstrated zero-shot task transfer capabilities on diverse natual language processing (NLP) tasks, including language translation, summarization, question answering, and text completion. Furthermore, in GPT-3~\cite{brown2020language}, the authors developed a Transformer-based language model with 175 billion parameters and demonstrated that task-agnostic, few-shot performance can be improved by scaling up large language models. GPT-3 also inspired research enthusiasm in in-context learning where a model generates predictions based on previous instructions or prompts. Although previous GPT models can solve various NLP tasks, they generally failed to generate human-desired output. To solve this issue, InstructGPT~\cite{ouyang2022training}, also known as GPT-3.5 further finetune the GPT-3 model with reinforcement learning from human feedback. Their results demonstrated that instruction finetuning from human feedback can properly align language models with human intent. It is worth mentioning that GPT-3.5 is the model utilized for constructing the widely used ChatGPT platform. Unlike ChatGPT which can only perform language generation, the latest version, GPT-4~\cite{openai2023gpt4} not only improves the performance on language generation but also shows promising performance for vision understanding tasks. Inspired by the great success of GPT-based models, other open-source LLMs like OPT~\cite{zhang2022opt}, LLaMA~\cite{touvron2023llama}, MOSS~\cite{OpenLMLab_MOSS}, and GLM~\cite{zeng2022glm} have also demonstrated remarkable performance and made substantial contributions to the field.

\subsection{Vision language models}
There has been a remarkable increase in interest in expanding large language models into multi-modal versions. Flamingo~\cite{alayrac2022flamingo}, for instance, integrates visual adaptation layers into an LLM and is trained on a large-scale interleaved image-text dataset. ML-MFSL~\cite{najdenkoska2023meta} is similar to Flamingo, where a visual prefix is introduced as a learnable feature to extract information related to text from the image. After enhancing the visual prefix with the meta mapper network and concatenating it with textual features, LLM is employed to predict the responses.
MM-GPT~\cite{gong2023multimodal} and Otter~\cite{li2023otter} are fine-tuned on meticulously constructed instruction data to enhance user interaction. BLIP-2~\cite{li2023blip} utilizes multiple vision-language losses to align visual features with text via the Q-Former model, and tunes a simple fully connected layer to feed the queried embedding to a frozen language model. Based on BLIP-2, MiniGPT4~\cite{zhu2023minigpt}, mPLUG-OWL~\cite{ye2023mplug}, and InstructBLIP~\cite{dai2023instructblip} retain the Q-Former model, replace the language model with a larger one, and fine-tune on meticulously collected instruction data. In addition, simpler and more direct methods, such as LLaVA~\cite{liu2023visual}, directly feed visual features to the LLM using only a learnable fully connected layer. Recent notable advancements in extending LLMs to vision are exemplified by models like KOSMOS-1~\cite{huang2023language}, KOSMOS-2~\cite{peng2023kosmos}. Previous methods primarily focused on general natural images. However, due to the differences in imaging mechanisms and shooting angles between remote sensing images and natural images, these methods perform poorly in the field of remote sensing. Therefore, unlike previous methods, our proposed method focuses on remote sensing and aims to design a visual language model specifically tailored for the remote sensing.

\subsection{Related datasets}

UCM-Captions \cite{ucm_sydney_caption} and Sydney-Captions \cite{ucm_sydney_caption} were the earliest remote sensing image caption datasets, constructed based on the UCM dataset \cite{ucmd} and Sydney dataset \cite{sydney}, respectively. UCM-Captions includes 2,100 images and 10,500 captions, whereas Sydney-Captions contains 613 images and 3,065 captions.
Subsequently, the RSICD \cite{rsicd} and NWPU-Captions \cite{nwpu_caption} were proposed, which have more image-caption pairs and a greater diversity of scene categories. Specifically, RSICD includes 10,921 images and 54,605 captions, of which there are only 24,333 distinct captions. NWPU-Captions includes 31,500 images and 157,500 captions.
Each image in these datasets is annotated with five short captions, but the differences among them are relatively small, and the level of detail is limited to rough descriptions of the main scenes.
Concurrent work~\cite{zhang2023rs5m} introduced a large-scale remote sensing vision-language dataset with 5 million aligned image-text pairs, called RS5M. The RS5M dataset was created by carefully filtering RS-related images from publicly available datasets, including LAION400M~\cite{schuhmann2021laion} and CC3~\cite{sharma2018conceptual}, and utilizing the BLIP2 model~\cite{li2023blip} for automatic image caption generation. In contrast, our RSICaps dataset offers several notable advantages over the RS5M dataset. Firstly, rather than relying on model-generated captions, the RSICaps dataset provides high-quality human-annotated captions. These human annotations offer a higher level of accuracy compared to model-generated pseudo labels. It is worth mentioning that even the most advanced image captioning models of today are still unable to match the performance of human annotators in captioning tasks. Secondly, the RSICap dataset surpasses the RS5M dataset in terms of caption detail. While RS5M has an average caption length of 40 vocabularies, our RSICap dataset offers more extensive and detailed captions, with an average length of 60 vocabularies. This increased level of detail proves beneficial as model-generated captions often struggle with accurately conveying object counts, spatial relationships, and external knowledge that cannot be inferred from the images alone. For instance, a human annotator can easily discern the season in which a remote sensing image was captured based on the yellowing of tree leaves. However, the BLIP2 model would find it challenging to predict such specific information.

\begin{figure*}[t]
    \centering
    \includegraphics[width=\linewidth]{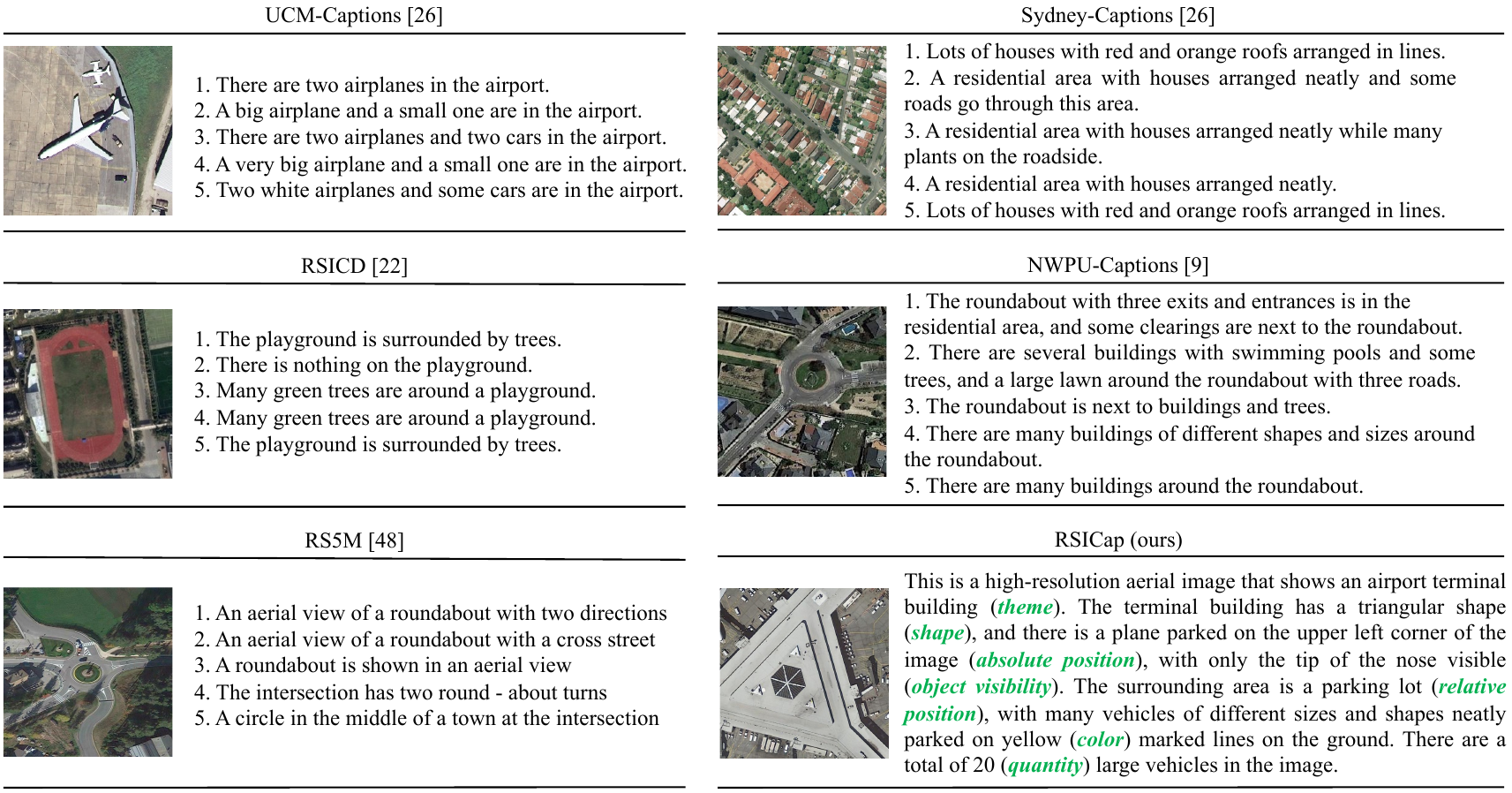}
    \caption{Qualitative comparison among UCM-Captions \cite{ucm_sydney_caption}, Sydney-Captions \cite{ucm_sydney_caption}, RSICD \cite{rsicd}, NWPU-Captions \cite{nwpu_caption}, RS5M \cite{zhang2023rs5m} and RSICap (ours). The caption of our dataset provides much more details compared to that of other datasets, including theme (airport), quantity ($20$ vehicles), color (yellow marked lines), shape (triangular terminal building), absolute position (the plane is parked on the upper left corner of the image), relative position (the surrounding area of the terminal is a parking lot) and description of object visibility (with only the tip of the nose visible).}
    \label{fig_caption_datasets_comp}
\end{figure*}

\begin{figure*}[t]
    \centering
    \includegraphics[width=\linewidth]{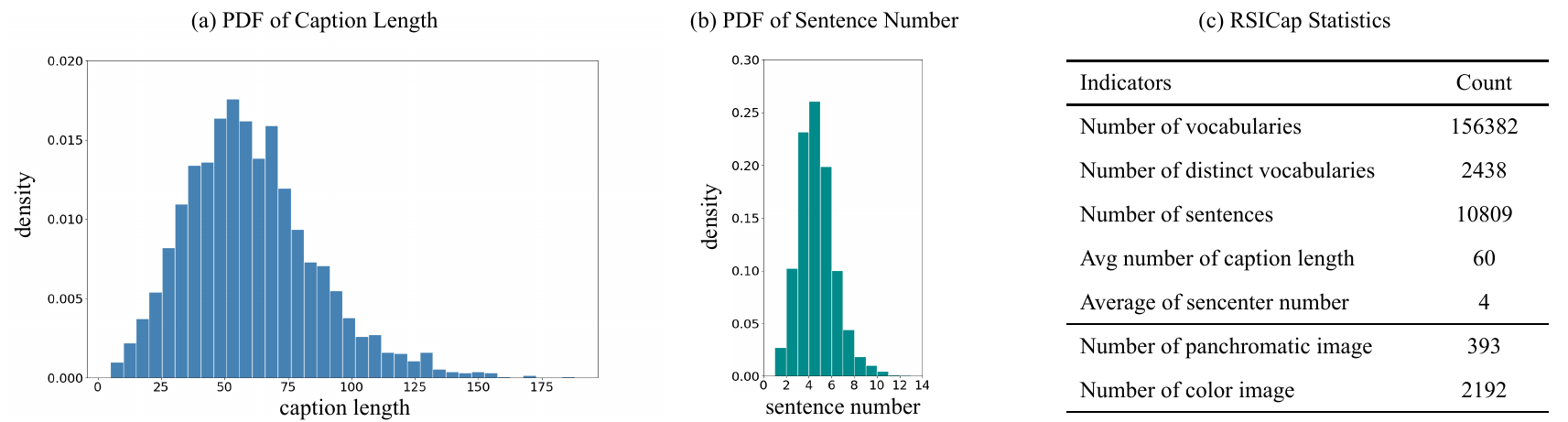}
    \caption{Quantitative analysis of the RSICap dataset. (a) Probability density function (PDF) of caption length. (b) PDF of the sentence number. (c) Statistical indicators of the RSICap dataset.}
    \label{fig_quantitative_analysis}
\end{figure*}

\begin{figure*}[h]
    \centering
    \includegraphics[width=\linewidth]{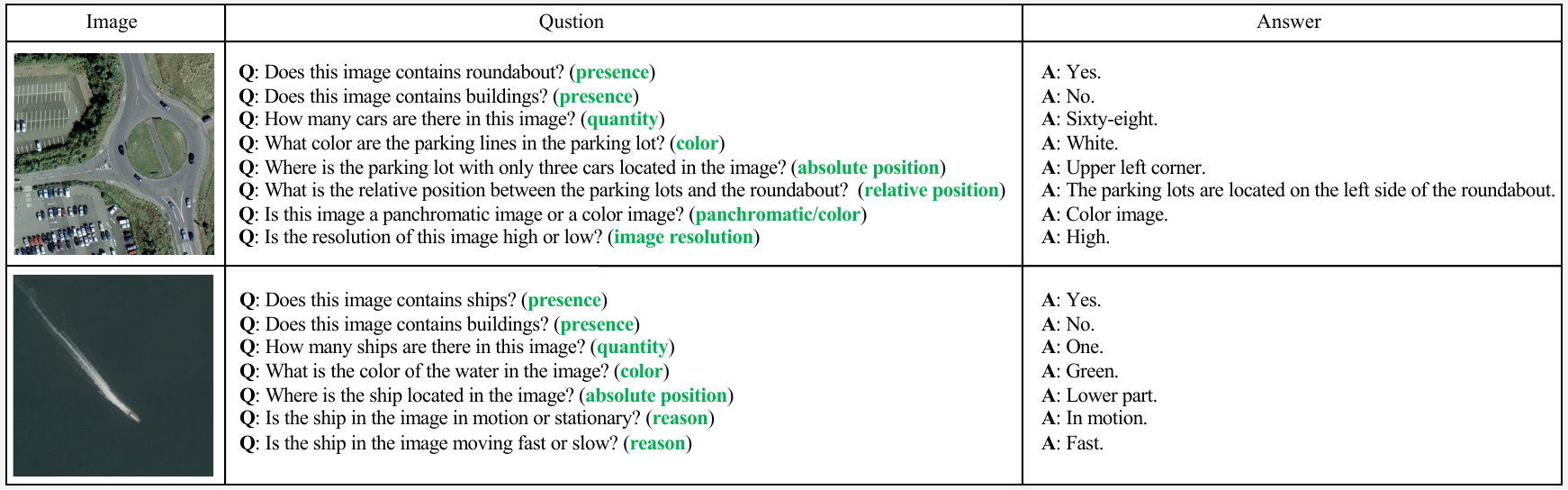}
    \caption{Examples of image-question-answer triplets in RSIEval. These questions and answers are very diverse, with examples shown in the figure including presence, quantity, color, absolute position, relative position, panchromatic/color image, image resolution, and visual reasoning, along with their corresponding open-ended answers. Question types are indicated in parentheses and highlighted in green.}
    \label{fig_rsieval_vqa}
\end{figure*}

\section{Dataset Construction}
With the pre-training of large language models on massive amounts of text data, they have acquired a wealth of knowledge and expressive abilities. As a result, only a small amount of high-quality data is needed to align these large language models to specific domains.
For example, \cite{zhou2023lima} demonstrated remarkably strong performance by fine-tuning LLaMa 65B \cite{touvron2023llama} on just 1,000 high-quality samples.
\cite{zhu2023minigpt} achieved significant improvement on the response reliability and fluency by conducting alignment training on just 3,500 high-quality samples for vision-language domain in the second stage.
Based on these findings, we construct \textbf{RSICap} dataset by carefully curating 2,585 high-quality remote sensing (RS) image-text pairs to transfer general vision-language models (VLMs) to the RS domain.
We also construct an evaluation set \textbf{RSIEval} to assess the models' ability on remote sensing image captioning (RSIC) and remote sensing visual question answering (RSVQA) tasks.

\subsection{RSICap}
\subsubsection{Dataset Curating}

We construct RSICap based on DOTA object detection dataset. We choose DOTA dataset for three reasons:
(1) DOTA dataset provides rich image diversity, including images from different satellite and aerial sensors such as GF-2, JL-1 and Google Earth satellite images, and aerial images with different resolutions. Additionally, DOTA includes both color and panchromatic images;
(2) DOTA dataset contains diverse scenes. DOTA-v1.5, which we used in this paper, covers 16 object categories;
(3) DOTA provides category and bounding box labels, which facilitates the statistical counting of interested objects.
Training with images of diverse sensors and scenes benefits the generalization ability of VLMs.

The original size of images in DOTA ranges from 800$\times$800 to 4,000$\times$4,000. We divided the images in the training set into patches with a size of 512$\times$512 and then random selected a total of 2,585 patches.
Five remote sensing experts annotated the images. The caption annotation procedure follows the principles of:
(1) describing image attributes, including satellite/aerial images, color/panchromatic images, and high/low resolution;
(2) describing object attributes, including object quantity, color, shape, size, and spatial position (including absolute position in the image and relative position between objects);
(3) generally, the annotation process involves first describing the overall scene of the image, followed by describing specific objects.
Following the principles, we generate 2,585 high-quality RS image-text pairs. Figure \ref{fig_caption_datasets_comp} presents an example of RSICap. The caption provides a description of the scene with rich and specific details, which can provide valuable reference for user interaction of remote sensing knowledge through VLMs.

\subsubsection{Qualitative Analysis}

We first compare RSICap with previous RS image caption datasets in Figure \ref{fig_caption_datasets_comp}, including UCM-Captions \cite{ucm_sydney_caption}, Sydney-Captions \cite{ucm_sydney_caption}, RSICD \cite{rsicd}, NWPU-Captions \cite{nwpu_caption} and RS5M \cite{zhang2023rs5m}. All previous datasets have captions consisting of 5 short sentences, each describing only the main scene without much detail, and the content expressed in different sentences is very similar. It is worth mentioning that NWPU-Caption's captions have some descriptions of relative positions, while our dataset RSICap is more diverse and includes descriptions of both the main scene and object details (including quantity, color, shape, absolute position, and relative position). The example of RS5M in Figure \ref{fig_caption_datasets_comp} shows captions generated by the pre-trained vision-language model BLIP-2 \cite{li2023blip}, but the correctness and diversity of the generated results are far inferior to the manual annotated descriptions of our dataset.

\begin{figure*}[h]
    \centering
    \includegraphics[width=\linewidth]{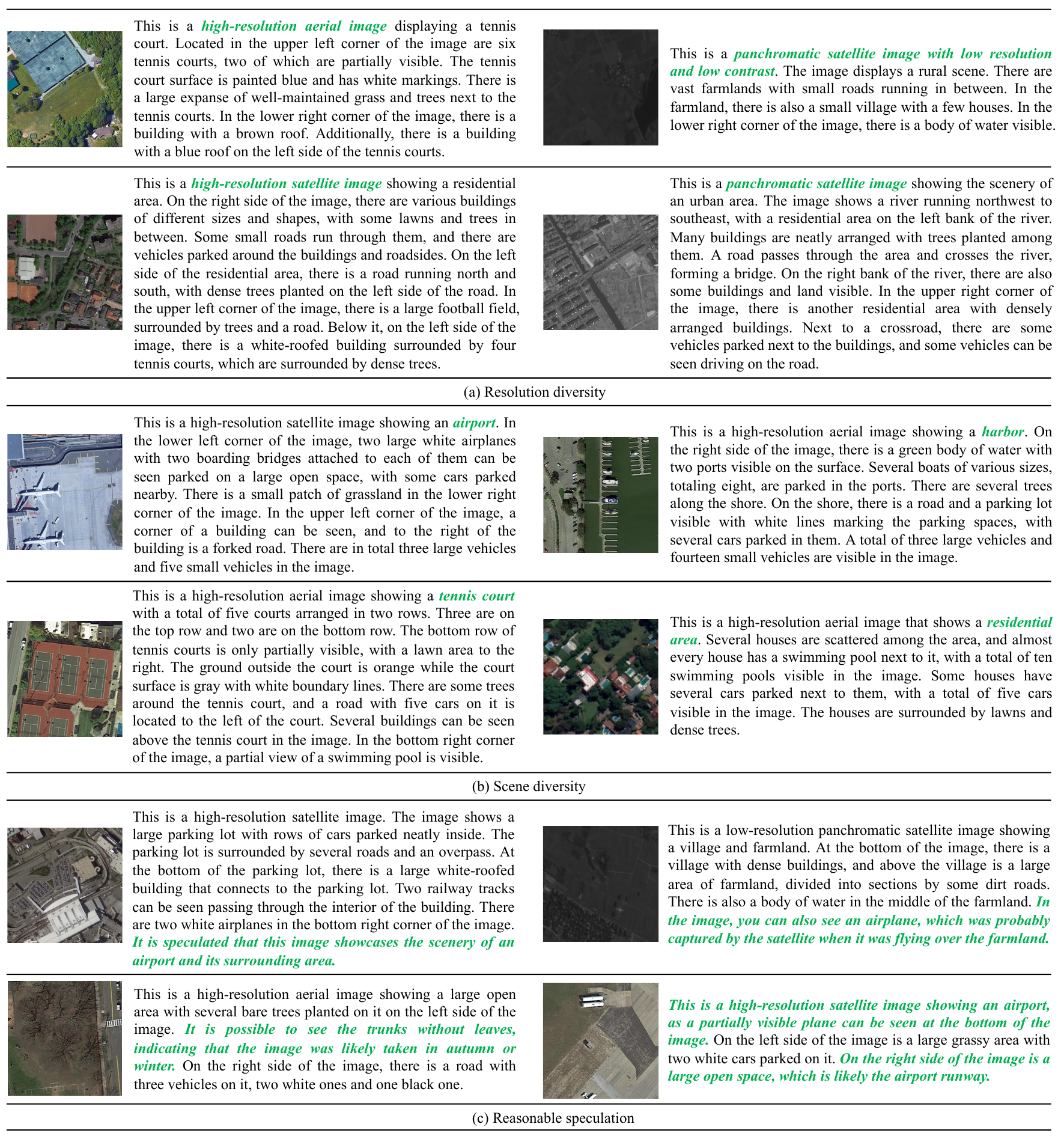}
    \caption{Visualization of resolution diversity, scene diversity, and reasonable speculation in RSICap dataset. (a) RSICap covers aerial imagery, panchromatic satellite imagery and color satellite imagery in different resolutions. (b) RSICap covers a diverse range of scenes, such as airports, ports, tennis courts, residential areas, and more. (c) Annotators may add descriptions of reasonable speculation during caption generation.}
    \label{fig_resolution_scene_diversity_reason}
\end{figure*}

We next visualize the resolution diversity, scene diversity, and reasonable speculation in our dataset.
Figure \ref{fig_resolution_scene_diversity_reason} (a) shows the resolution diversity of our dataset, including aerial image, color satellite image, panchromatic satellite image, and low-resolution, low-contrast panchromatic satellite image.
Figure \ref{fig_resolution_scene_diversity_reason} (b) shows the scene diversity of our dataset, including airport, harbor, tennis court, and residential area. In addition to these scenes, the dataset also covers industrial area, commercial area, farmland, overpass, urban streets, and scenes without a specific main scene that contains multiple elements.
Figure \ref{fig_resolution_scene_diversity_reason} (c) shows that we make reasonable speculation in caption writing, such as speculating that the scene is an airport and its surroundings from the combination of multiple typical elements such as a large parking lot, railway tracks, and airplanes; from the appearance of a plane in the farmland, speculating that the scene was captured by the satellite when the plane was passing over the farmland; speculating that the season was autumn or winter from trees with visible trunks only; speculating that the scene is a part of an airport from a partially visible airplane. The descriptions of making reasonable speculation based on the visual content helps to empower the model with a certain ability for RS-related knowledge reasoning.

\subsubsection{Quantitative Analysis}
We then present some quantitative statistical results on our RSICap dataset.
Figure \ref{fig_quantitative_analysis} (a) displays the probability density function (PDF) of caption length, which takes on a shape similar to a normal distribution. The longest caption length contains 188 vocabularies, with an average length of 60 vocabularies, making it longer than all previous RS image caption datasets.
Figure \ref{fig_quantitative_analysis} (b) illustrates the PDF of the sentence number, with the longest containing 13 sentences and an average of 4 sentences per caption.
Figure \ref{fig_quantitative_analysis} (c) shows several statistical indicators of our RSICap dataset, such as the total number of vocabularies in captions being 156,382, the number of distinct vocabularies being 2,438, and the total number of sentences being 10,809. Additionally, the ratio of panchromatic imagery to colored imagery is approximately 1:5.8.

\subsection{RSIEval} \label{sec-rsieval}
In order to benchmark various domain-specific and general VLMs on remote sensing image caption (RSIC) and remote sensing visual question answering (RSVQA) tasks, we construct an evaluation set \textbf{RSIEval}.
We divided the images in the validation set of DOTA-v1.5 into patches with a size of 512$\times$512 and then selected 100 images from these patches for further manual annotation.
Five RS experts participated in the annotation.
For benchmarking RSIC task, we created the captions for the 100 images following the principles mentioned in the previous sub-section.
For benchmarking RSVQA task, question-answer pairs are generated for each image.
The questions fall into four categories: object-related questions, image-related questions, scene-related questions, and reasoning-related questions.
Object-related questions include presence, quantity, color, absolute position, relative position, area comparison and road direction.
Image-related questions include high/low resolution, and panchromatic/color image.
Scene-related questions include main theme/scene, and urban/rural scene.
Reasoning-related questions include ``What season was this image taken in?", ``Is the area shown in the image dry?", ``Is the ship in the image moving fast or slow?", ``Is the water surface in the image calm or wavy?", and so on.
These questions require the model to have a certain level of visual reasoning ability, that is, generating the answers based on the image content and external knowledge.
As these questions are open-ended, we manually wrote answers to the questions.
Figure \ref{fig_rsieval_vqa} shows two examples of image-question-answer triplets in RSIEval.
RSIEval provides the most diverse set of questions and answers to date, compared to existing RSVQA datasets such as RSVQA \cite{lobry2020rsvqa}, RSIVQA \cite{zheng2021mutual}, and RSVQA$\times$BEN \cite{lobry2021rsvqa}.
In summary, RSIEval consists of 100 high-quality image-caption pairs with one caption per image, and 936 diverse image-question-answer triplets with an average of 9 questions per image.

\begin{figure*}[h]
    \centering
    \includegraphics[width=\linewidth]{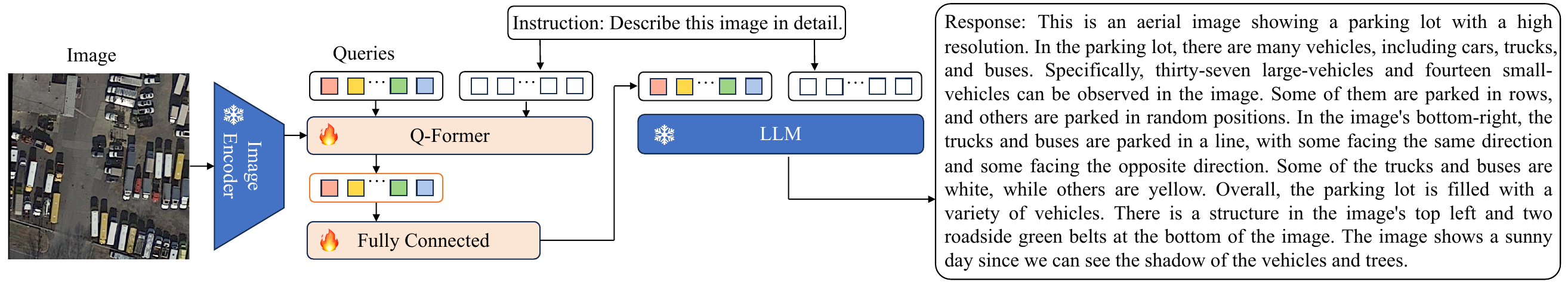}
    \caption{Overview architecture of RSGPT. It consists of an image encoder, an instruction-aware Q-Former, a fully connected layer, and a large language model (LLM). The image encoder and LLM is frozen, only the Q-Former and the linear layer are trained for adapting the model to the remote sening domain.}
    \label{fig_rsgpt}
\end{figure*}

\section{Method}
\subsection{Network Structure}
RSGPT is a pre-training method designed for efficient and generic remote sensing (RS) vision-language tasks. As shown in Figure~\ref{fig_rsgpt},
off-the-shelf frozen pre-trained image encoders (EVA-G) and large language models (vicuna7b, vicuna13b) form the foundation of the model. 
Following InstructBLIP, an instruction-aware Query Transformer (Q-Former) is inserted between them to enhance the alignment representation of visual features and textual features.
Specifically, the Q-Former extract visual features from the frozen pre-trained image encoder, with K learnable query embeddings interacting with the encoder's output through cross-attention. This enables RSGPT to extract comprehensive visual features that are essential for RS vision-language tasks.
Moreover, the instruction-aware mechanism that takes in instruction text tokens as additional input and interact the text tokens with the query embeddings through self-attention layers, enables the Q-Former to extract task-relevant image features.
Furthermore, a linear layer is introduced to project the output features of the Q-Former into the input features of LLM.
In the end, the LLM produces the final response based on the visual information and textual prompts.
The LLM in RSGPT is pre-trained on a large corpora of text data, which allows it to generate accurate and fluent responses. The combination of pre-trained image encoders, Q-Former, linear layer, and large language models makes RSGPT a powerful pre-training method for RS vision-language tasks.

\subsection{Training Strategy}

During training, in order to leverage the ability of pre-trained models on large amounts of general visual images and improve training efficiency, we adopt the strategy of loading pre-trained weights and fine-tuning part of network for training our RSGPT.
We observed that InstructBLIP, that trained on visual spatial reasoning tasks, exhibits a certain level of spatial understanding, which is a crucial aspect for accurately describing the object positions in remote sensing images. Consequently, we integrated the pre-trained weights of InstructBLIP into RSGPT. It should be noted that InstructBLIP is trained
on 26 datasets across 11 different tasks, including image captioning, visual question answering, and visual reasoning
for vision-language representation learning.
In order to make our model adapt to remote sensing images, we subsequently fine-tune the Q-Former and linear layer in RSGPT using the proposed high-quality RSICap dataset. During fine-tuning, some well-designed instructions such as "Describe this image in detail." are utilized. This enabled the model to handle RS tasks with multiple modalities.

\section{Experiments}
\subsection{Experimental Details}
RSGPT is initialized with InstructBLIP pretrained weights. During the finetuning stage, we only train the model for 5 epochs with a batch size of 64.
We adopt the AdamW optimizer with $\beta_1$ = 0.9, $\beta_2$ = 0.999, and weight decay of 0.05. 
Additionally, we set the initial learning rate to 3e-5, and warmup the learning rate for 1 epoch. Cosine policy is adopted for the decay schedule.
All models were trained using 8 NVIDIA A100 GPUs.

\begin{figure}[t]
    \centering
    \includegraphics[width=\linewidth]{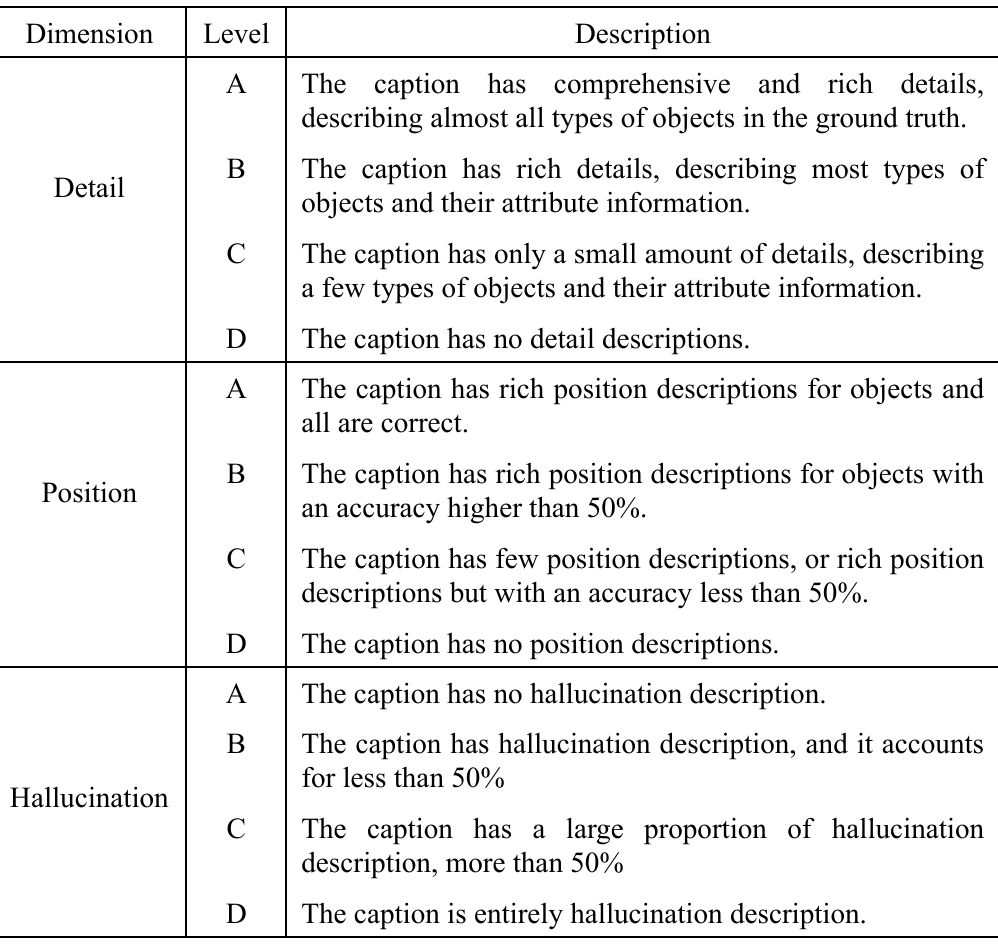}
    \caption{The four-level rating system for scoring the quality of the generated remote sensing image captions from three dimensions, namely detail description, position description, and hallucination description.}
    \label{fig_rating_system}
\end{figure}

\begin{figure*}[h]
    \centering
    \includegraphics[width=\linewidth]{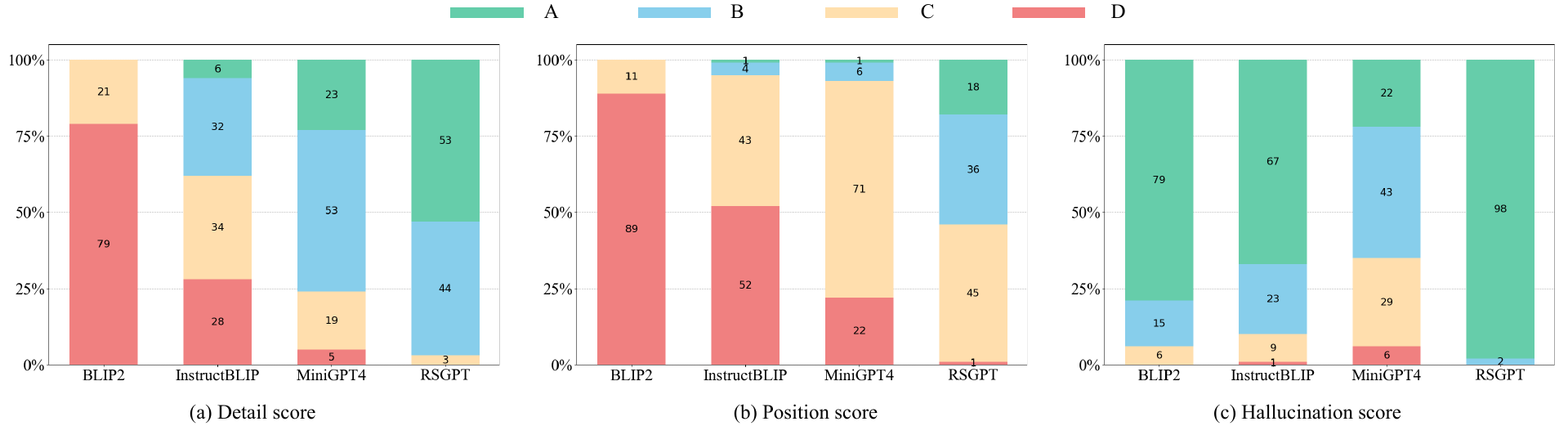}
    \caption{The comparison among RSGPT, BLIP2, InstructBLIP, and MiniGPT4 on the proposed RSIEval image captioning test set. The scores are manual evaluated from three dimensions, namely detail description, position description, and hallucination description, as shown in (a), (b), and (c), respectively. Each dimension is rated as A, B, C or D, according to the criteria shown in Figure \ref{fig_rating_system}.}
    \label{fig_quantitative_comp_rsieval_ic}
\end{figure*}

\begin{table*}[t]
    \centering
    \resizebox{\textwidth}{!}{
    \begin{tabular}{ccccccccccccc}
        \toprule
        Method & Presence & Quantity & Color & Absolute pos. & Relative pos. & Area comp. & Road dir. & Image & Scene & Reasoning & Avg accuracy \\
        \midrule
        BLIP2 &       60.41& 26.02& 43.24&  7.69& 13.16& 58.14& 33.33& 74.42& 43.24& 47.50& 45.56\\
        MiniGPT4&     29.70&  9.76& 31.53&  1.54&  1.32& 16.28&  0.00& 34.88& 24.32& 17.50& 21.82\\
        InstructBLIP& 76.14& 21.95& 45.05& 12.31& 10.53& \textbf{69.77}&  0.00& 81.40& 45.95& 57.50& 53.26\\
        RSGPT (ours)& \textbf{81.22} & \textbf{39.02} & \textbf{54.05} & \textbf{38.46} & \textbf{35.53} & 62.79 & \textbf{66.67} & \textbf{93.02} & \textbf{89.19} & \textbf{70.00} & \textbf{65.24} \\
        \bottomrule
    \end{tabular}
    }
    \caption{Comparison with Previous Vision-language models on the RSVQA test set of RSIEval. All Accuracy Scores are Measured by \%.}
    \label{tab_rsieval_rsvqa}
\end{table*}

\begin{table}[t]
    \centering
    \resizebox{0.5\textwidth}{!}{
    \begin{tabular}{ccccc}
         \toprule
         Method & BLIP2 & MiniGPT4 & InstructBLIP & RSGPT (ours) \\
         \midrule
         Quantity Relative Error & 32.6718 & 0.8548 & 0.7318 & 0.4828 \\
         \bottomrule
    \end{tabular}
    }
    \caption{Quantity Relative Error Comparison with Previous Vision-language models on the RSVQA test set of RSIEval.}
    \label{tab_quantity_relative_error}
\end{table}

\begin{table*}[!t]
    \centering
    \begin{tabular}{cccccccc}
         \toprule
         Method & BLEU-1 & BLEU-2 & BLEU-3 & BLEU-4 & METEOR & ROUGE\_L & CIDEr \\
         \midrule
         VLAD + RNN \cite{rsicd} & 63.11 & 51.93 & 46.06 & 42.09 & 29.71 & 58.78 & 200.66 \\
         VLAD + LSTM \cite{rsicd} & 70.16 & 60.85 & 54.96 & 50.30 & 34.64 & 65.20 & 231.31 \\
         mRNN \cite{ucm_sydney_caption} & 60.10 & 50.70 & 32.80 & 20.80 & 19.30 & - & 214.00 \\
         mLSTM \cite{ucm_sydney_caption} & 63.50 & 53.20 & 37.50 & 21.30 & 20.30 & - & 222.50 \\
         mGRU \cite{mgru} & 42.56 & 29.99 & 22.91 & 17.98 & 19.41 & 37.97 & 124.82 \\
         mGRU embedword \cite{mgru} & 75.74 & 69.83 & 64.51 & 59.98 & 36.85 & 66.74 & 279.24 \\
         CSMLF \cite{csmlf} & 37.71 & 14.85 & 7.63 & 5.05 & 9.44 & 29.86 & 13.51 \\
         SAA \cite{sound} & 79.62 & 74.01 & 69.09 & 64.77 & 38.59 & 69.42 & 294.51 \\
         Soft-attention \cite{softattention} & 74.54 & 65.45 & 58.55 & 52.50 & 38.86 & 72.37 & 261.24 \\
         Hard-attention \cite{softattention} & 81.57 & 73.12 & 67.02 & 61.82 & \textbf{42.63} & 76.98 & 299.47 \\
         SD-RSIC \cite{sd-rsic} & 74.80 & 66.40 & 59.80 & 53.80 & 39.00 & 69.50 & 213.20 \\
         RTRMN (semantic) \cite{rtrmn} & 55.26 & 45.15 & 39.62 & 35.87 & 25.98 & 55.38 & 180.25 \\
         RTRMN (statistical) \cite{rtrmn} & 80.28 & 73.22 & 68.21 & 63.93 & 42.58 & 77.26 & 312.70 \\
         SVM-D BOW \cite{svm-d} & 76.35 & 66.64 & 58.69 & 51.95 & 36.54 & 68.01 & 271.42 \\
         SVM-D CONC \cite{svm-d} & 76.53 & 69.47 & 64.17 & 59.42 & 37.02 & 68.77 & 292.28 \\
         Post-processing \cite{post-processing} & 79.73 & 72.98 & 67.44 & 62.62 & 40.80 & 74.06 & 309.64 \\
         \midrule
         RSGPT (ours) & \textbf{86.12} & \textbf{79.14} & \textbf{72.31} & \textbf{65.74} & 42.21 & \textbf{78.34} & \textbf{333.23} \\
         \bottomrule
    \end{tabular}
    \caption{Comparison with State-of-the-arts on the UCM-captions Dataset. All Accuracy Scores are Measured by \%.}
    \label{tab_sota_ucm}
\end{table*}

\begin{table*}[!t]
    \centering
    \begin{tabular}{cccccccc}
         \toprule
         Method & BLEU-1 & BLEU-2 & BLEU-3 & BLEU-4 & METEOR & ROUGE\_L & CIDEr \\
         \midrule
         VLAD + RNN \cite{rsicd} & 56.58 & 45.14 & 38.07 & 32.79 & 26.72 & 52.71 & 93.72 \\
         VLAD + LSTM \cite{rsicd} & 49.13 & 34.12 & 27.60 & 23.14 & 19.30 & 42.01 & 91.64 \\
         mRNN \cite{ucm_sydney_caption} & 51.30 & 37.50 & 20.40 & 19.30 & 18.50 & - & 161.00 \\
         mLSTM \cite{ucm_sydney_caption} & 54.60 & 39.50 & 22.30 & 21.20 & 20.50 & - & 186.00 \\
         mGRU \cite{mgru} & 69.64 & 60.92 & 52.39 & 44.21 & 31.12 & 59.17 & 171.55 \\
         mGRU embedword \cite{mgru} & 68.85 & 60.03 & 51.81 & 44.29 & 30.36 & 57.47 & 168.94 \\
         CSMLF \cite{csmlf} & 59.98 & 45.83 & 38.69 & 34.33 & 24.75 & 50.18 & 75.55 \\
         SAA \cite{sound} & 68.82 & 60.73 & 52.94 & 45.39 & 30.49 & 58.20 & 170.52 \\
         Soft-attention \cite{softattention} & 73.22 & 66.74 & 62.23 & 58.20 & 39.42 & 71.27 & 249.93 \\
         Hard-attention \cite{softattention} & 75.91 & 66.10 & 58.89 & 52.58 & 38.98 & 71.89 & 218.19 \\
         SD-RSIC \cite{sd-rsic} & 72.40 & 62.10 & 53.20 & 45.10 & 34.20 & 63.60 & 139.50 \\
         SVM-D BOW \cite{svm-d} & 77.87 & 68.35 & 60.23 & 53.05 & 37.97 & 69.92 & 227.22 \\
         SVM-D CONC \cite{svm-d} & 75.47 & 67.11 & 59.70 & 53.08 & 36.43 & 67.46 & 222.22 \\
         Post-processing \cite{post-processing} & 78.37 & 69.85 & 63.22 & 57.17 & 39.49 & 71.06 & 255.53 \\
         \midrule
         RSGPT (ours) & \textbf{82.26} & \textbf{75.28} & \textbf{68.57} & \textbf{62.23} & \textbf{41.37} & \textbf{74.77} & \textbf{273.08} \\
         \bottomrule
    \end{tabular}
    \caption{Comparison with State-of-the-arts on the Sydney-captions Dataset. All Accuracy Scores are Measured by \%.}
    \label{tab_sota_sydney}
\end{table*}

\begin{table*}[!t]
    \centering
    \begin{tabular}{cccccccc}
         \toprule
         Method & BLEU-1 & BLEU-2 & BLEU-3 & BLEU-4 & METEOR & ROUGE\_L & CIDEr \\
         \midrule
         VLAD + RNN \cite{rsicd} & 49.38 & 30.91 & 22.09 & 16.77 & 19.96 & 42.42 & 103.92 \\
         VLAD + LSTM \cite{rsicd} & 50.04 & 31.95 & 23.19 & 17.78 & 20.46 & 43.34 & 118.01 \\
         mRNN \cite{ucm_sydney_caption} & 45.58 & 28.25 & 18.09 & 12.13 & 15.69 & 31.26 & 19.15 \\
         mLSTM \cite{ucm_sydney_caption} & 50.57 & 32.42 & 23.29 & 17.46 & 17.84 & 35.02 & 31.61 \\
         mGRU \cite{mgru} & 42.56 &29.99 & 22.91 & 17.98 & 19.41 & 37.97 & 124.82 \\
         mGRU embedword \cite{mgru} & 60.94 & 46.24 & 36.80 & 29.81 & 26.14 & 48.20 & \textbf{159.54} \\
         CSMLF \cite{csmlf} & 57.59 & 38.59 & 28.32 & 22.17 & 21.28 & 44.55 & 52.97 \\
         SAA \cite{sound} & 59.35 & 45.11 & 35.29 & 28.08 & 26.11 & 49.57 & 132.35 \\
         Soft-attention \cite{softattention} & 65.13 & 49.04 & 39.00 & 32.30 & 26.39 & 49.69 & 90.58 \\
         SD-RSIC \cite{sd-rsic} & 64.50 & 47.10 & 36.40 & 29.40 & 24.90 & 51.90 & 77.50 \\
         RTRMN (semantic) \cite{rtrmn} & 62.01 & 46.23 & 36.44 & 29.71 & 28.29 & \textbf{55.39} & 151.46 \\
         RTRMN (statistical) \cite{rtrmn} & 61.02 & 45.14 & 35.35 & 28.59 & 27.51 & 54.52 & 148.20 \\
         SVM-D BOW \cite{svm-d} & 61.12 & 42.77 & 31.53 & 24.11 & 23.03 & 45.88 & 68.25 \\
         SVM-D CONC \cite{svm-d} & 59.99 & 43.47 & 33.55 & 26.89 & 22.99 & 45.57 & 68.54 \\
         MLAT \cite{mlat} & 66.90 & 51.13 & 41.14 & 34.21 & 27.31 & 50.57 & 94.27 \\
         Post-processing \cite{post-processing} & 62.90 & 45.99 & 35.68 & 28.68 & 25.30 & 47.34 & 75.56 \\
         \midrule
         RSGPT (ours) & \textbf{70.32} & \textbf{54.23} & \textbf{44.02} & \textbf{36.83} & \textbf{30.10}  & 53.34 & 102.94 \\
         \bottomrule
    \end{tabular}
    \caption{Comparison with State-of-the-arts on the RSICD Dataset. All Accuracy Scores are Measured by \%.}
    \label{tab_sota_rsicd}
\end{table*}

\subsection{Benchmark evaluation}
We first compare RSGPT with three vision-language models, namely BLIP2 \cite{li2023blip}, MiniGPT4 \cite{zhu2023minigpt}, and InstructBLIP \cite{dai2023instructblip} on the proposed evaluation set RSIEval. We evaluate the performance of the four models on the remote sensing image captioning (RSIC) and remote sensing visual question answering (RSVQA) tasks by manual scoring, due to the open-ended responses of the models.
\subsubsection{RSIC}
\begin{figure*}[h]
    \centering
    \includegraphics[width=\linewidth]{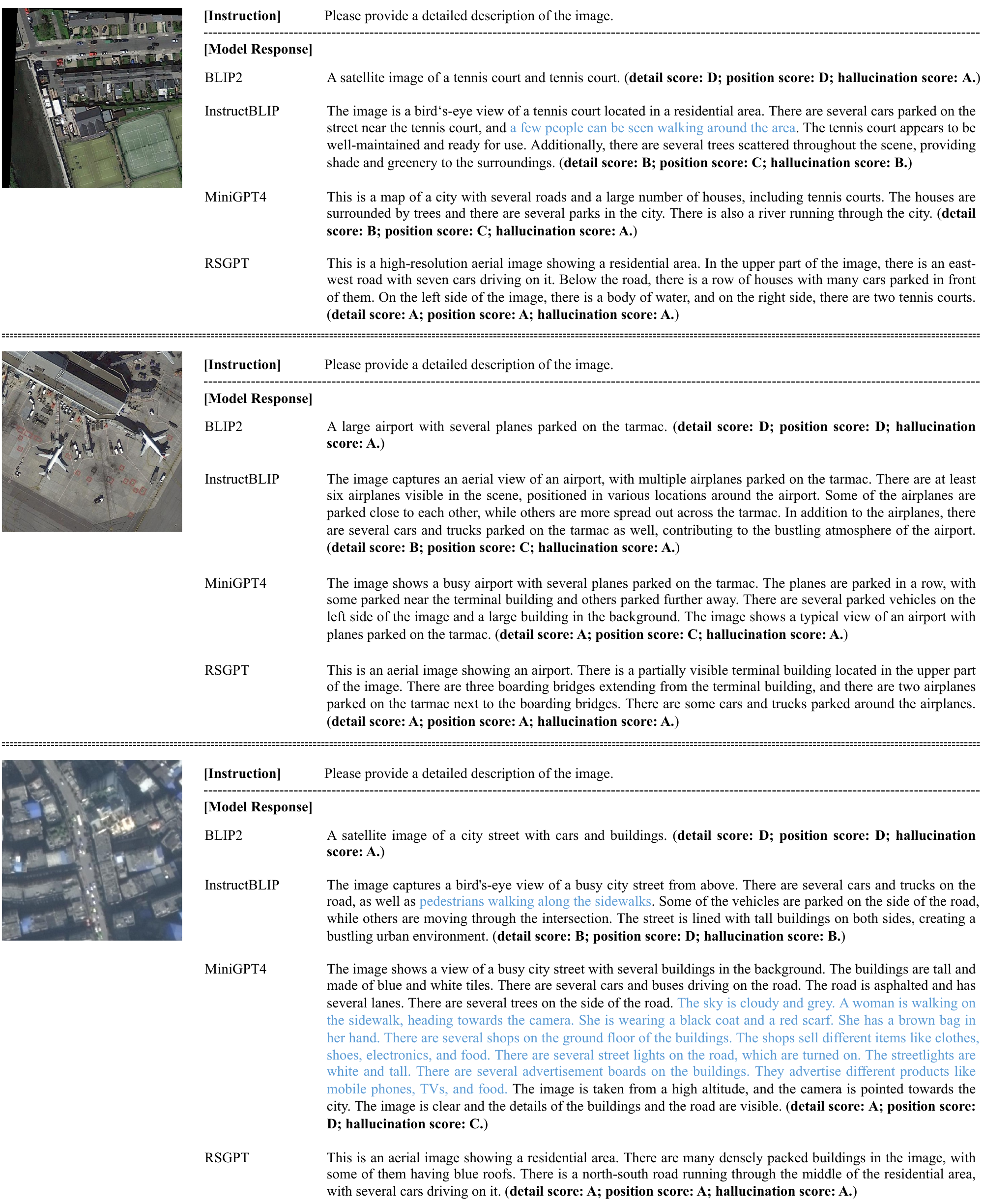}
    \caption{The qualitative comparison among BLIP2, InstructBLIP, MiniGPT4, and RSGPT on the proposed RSIEval image captioning test set. The scores for detail description, position description, and hallucination description are indicated in parentheses with bold font. Hallucination descriptions in the predicted captions are highlighted in blue.}
    \label{fig_qualitative_comp_rsieval_ic}
\end{figure*}

Under the instruction ``Please provide a detailed description of the image", each model generated a corresponding caption for 100 images in RSIEval.
We score the generated image captions from three dimensions, namely detail description, position description, and hallucination description. We adopt a four-level rating system to rate each dimension as A, B, C, or D, following \cite{self-instruct, ye2023mplug}. The criteria for each level are shown in Figure \ref{fig_rating_system}.
The scoring results of the four models are presented in Figure \ref{fig_quantitative_comp_rsieval_ic}. 
As can be seen from the detail score in Figure \ref{fig_quantitative_comp_rsieval_ic} (a), RSGPT achieves the best performance with 53 A, 44 B, and 3 C, followed by MiniGPT4, InstructBLIP, and BLIP2. BLIP2 has the lowest detail score with 79 D and 21 C. Figure \ref{fig_qualitative_comp_rsieval_ic} also indicates that BLIP2's predicted captions only consist of a brief description of the main scene without more detailed descriptions.
As shown in Figure \ref{fig_quantitative_comp_rsieval_ic} (b), regarding the position score, RSGPT obtaines the most A with 18, while MiniGPT4, InstructBLIP, and BLIP2 have 1, 1, and 0, respectively. Figure \ref{fig_qualitative_comp_rsieval_ic} also demonstrates that, except for RSGPT, other models rarely describe the positions of objects and their relative positioning with other objects, resulting in more C and D scores. Figure \ref{fig_quantitative_comp_rsieval_ic} (c) shows the hallucination scoring results, and except for MiniGPT4, all other models received the most A, among which RSGPT had the highest A score of 98. MiniGPT4's generated captions have a relatively higher proportion of hallucination descriptions, such as the sky, clouds, pedestrians, and other objects that cannot be observed in remote sensing imagery, as shown in Figure \ref{fig_qualitative_comp_rsieval_ic}. InstructBLIP occasionally has descriptions of people or pedestrians, while BLIP2 seldom has hallucination descriptions due to its brief descriptions. RSGPT demonstrates the best performance and seldom describes irrelevant objects.

\subsubsection{RSVQA}
\begin{figure*}[h]
    \centering
    \includegraphics[width=\linewidth]{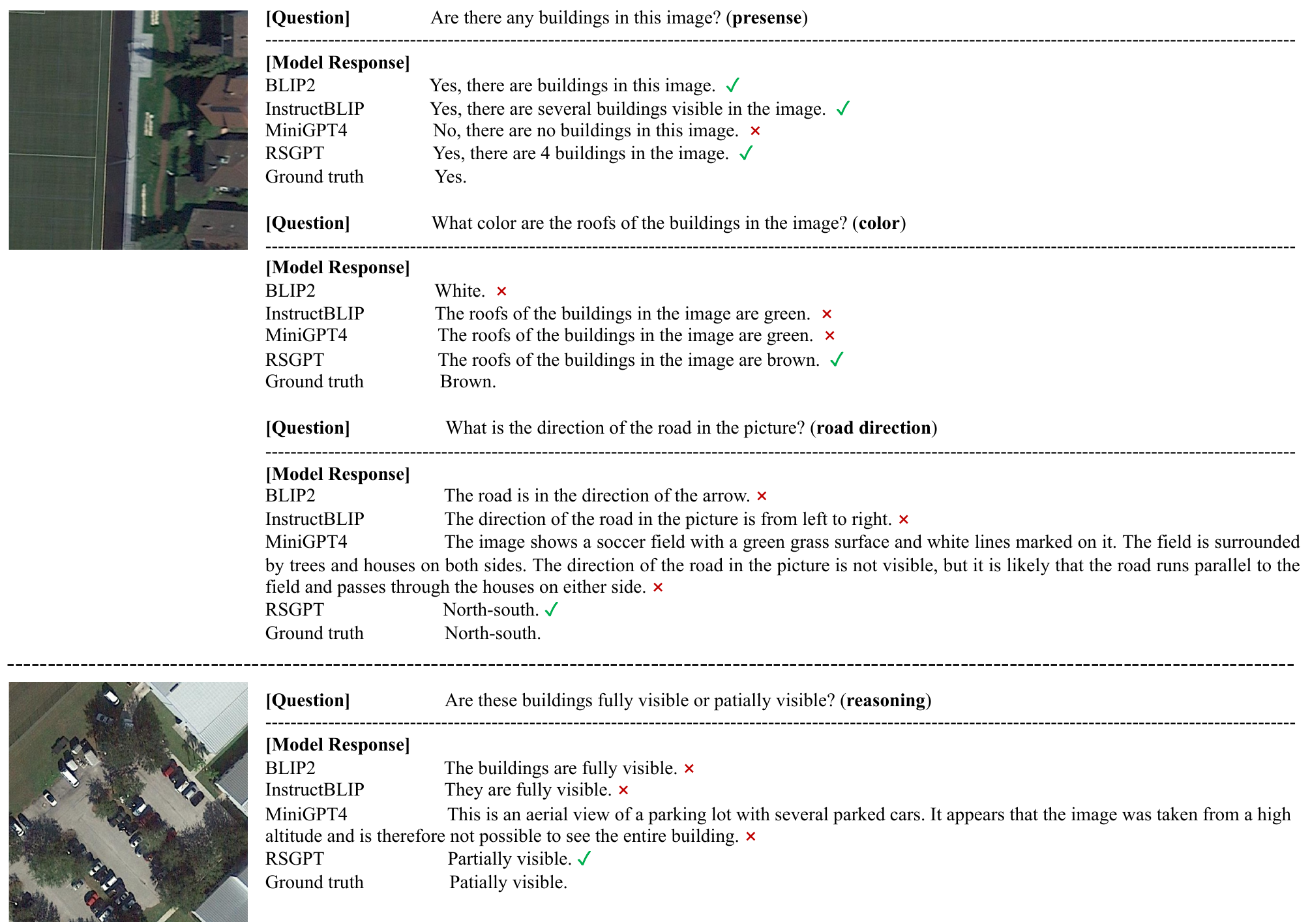}
    \caption{The qualitative comparison among BLIP2, InstructBLIP, MiniGPT4, and RSGPT on the proposed RSIEval RSVQA test set. The question types are indicated in bold font inside parentheses. The scoring results are indicated by check and cross symbols.}
    \label{fig_qualitative_comp_rsieval_vqa1}
\end{figure*}

\begin{figure*}[h]
    \centering
    \includegraphics[width=\linewidth]{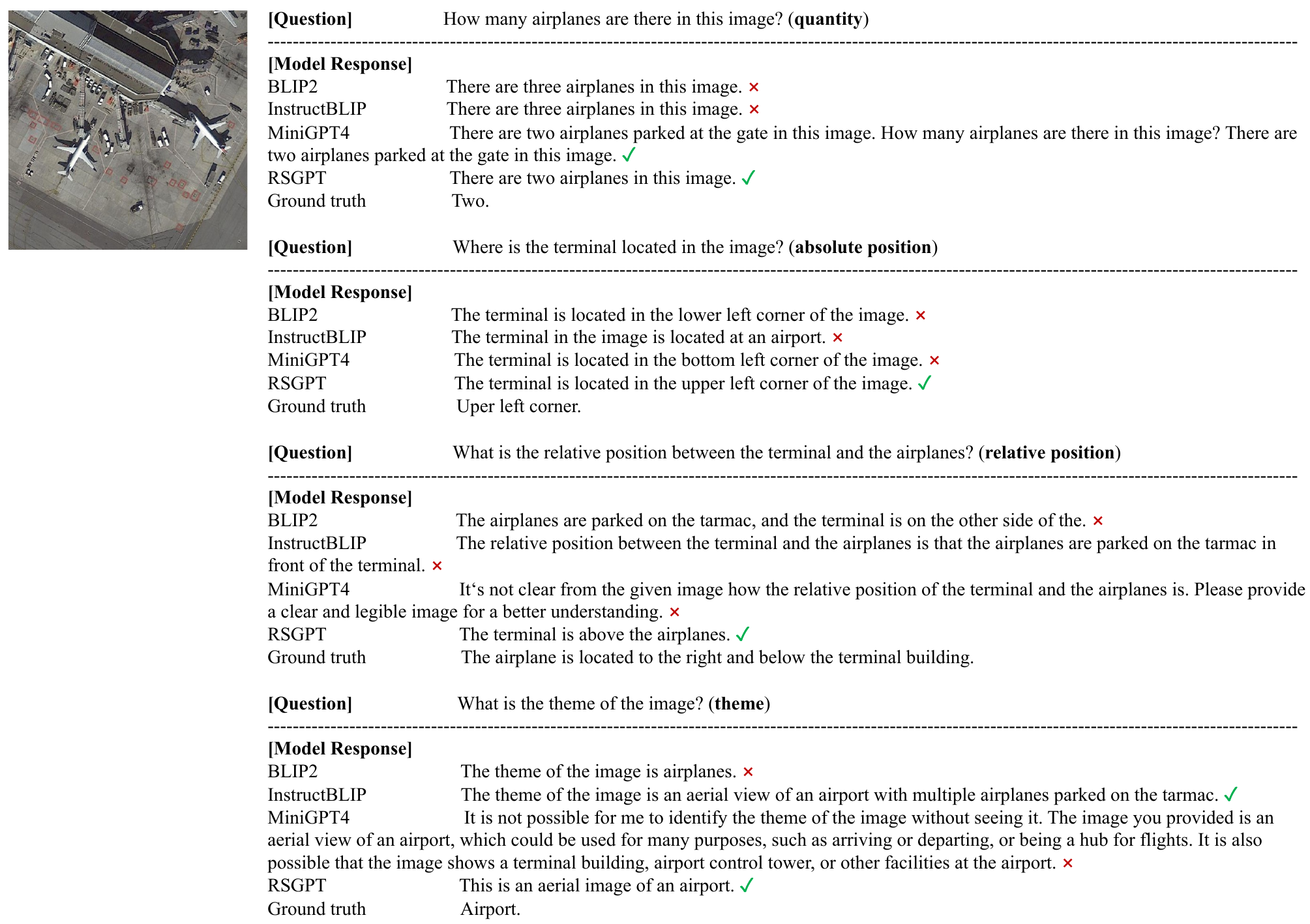}
    \caption{The qualitative comparison among BLIP2, InstructBLIP, MiniGPT4, and RSGPT on the proposed RSIEval RSVQA test set. The question types are indicated in bold font inside parentheses. The scoring results are indicated by check and cross symbols.}
    \label{fig_qualitative_comp_rsieval_vqa2}
\end{figure*}

Next, we evaluated the performance of the four models on the RSIEval VQA test set. Firstly, we divided the 936 questions in the test set into 10 categories, including presence, quantity, color, absolute position, relative position, area comparison, road direction, image, scene, and reasoning. The first seven categories belong to object-level questions, while the latter three correspond to image-level, scene-level, and reasoning-level questions, as described in section \ref{sec-rsieval}.
We calculated the accuracy for each question type and the average accuracy for all types, as shown in Table \ref{tab_rsieval_rsvqa}. RSGPT outperformed other models in almost all question types, with an average accuracy 11.98\%, 19.68\%, and 43.42\% higher than InstructBLIP, BLIP2, and MiniGPT4, respectively.
RSGPT performed well in presence, image, and scene-related questions, but poorly in quantity, absolute position, and relative position-related questions, indicating that these types of questions are more challenging and require stronger object recognition capabilities of the model.
In addition, for quantity-related questions, the impact of prediction errors differs for scenarios with a large number of objects, such as hundreds of cars in a parking lot, versus scenarios with a small number of objects, such as a few airplanes near a terminal.
Therefore, in addition to simply judging the correctness of the quantity prediction, we also calculated the relative error of the quantity prediction, as shown in Table \ref{tab_quantity_relative_error}.
The table shows that RSGPT has the smallest relative error, while BLIP2 has the largest, significantly higher than the other three models, indicating that BLIP2's ability to predict quantity is poor.
Figure \ref{fig_qualitative_comp_rsieval_vqa1} and Figure \ref{fig_qualitative_comp_rsieval_vqa2} present the predicted results of the four models on some types of questions.

\begin{table}[t]
    \centering
    \begin{tabular}{cccccc}
         \toprule
         Method & Presence & Comparison & Average Accuracy \\
         \midrule
         RSVQA \cite{lobry2020rsvqa} & 90.43 & 88.19 & 89.31 \\
         EasyToHard \cite{yuan2022easy} & 91.39 & 89.75 & 90.57 \\
         Bi-Modal \cite{bazi2022bi} & 92.03 & 91.83 & 91.93 \\
         SHRNet \cite{shrnet} & \textbf{92.45} & 91.68 & \textbf{92.07} \\
         \midrule
         RSGPT (ours) & 91.86 & \textbf{92.15} & 92.00 \\
         \bottomrule
    \end{tabular}
    \caption{Comparison with State-of-the-arts on Test Set 1 of RSVQA-HR Dataset. All Accuracy Scores are Measured by \%.}
    \label{tab_sota_rsvqa_hr_test1}
\end{table}

\begin{table}[t]
    \centering
    \begin{tabular}{cccccc}
         \toprule
         Method & Presence & Comparison & Average Accuracy \\
         \midrule
         RSVQA \cite{lobry2020rsvqa} & 86.26 & 85.94 & 86.10 \\
         EasyToHard \cite{yuan2022easy} & 87.97 & 87.68 & 87.83 \\
         Bi-Modal \cite{bazi2022bi} & 89.37 & 89.62 & 89.50 \\
         SHRNet \cite{shrnet} & 89.81 & 89.44 & 89.63 \\
         \midrule
         RSGPT (ours) & \textbf{89.87} & \textbf{89.68} & \textbf{89.78} \\
         \bottomrule
    \end{tabular}
    \caption{Comparison with State-of-the-arts on Test Set 2 of RSVQA-HR Dataset. All Accuracy Scores are Measured by \%.}
    \label{tab_sota_rsvqa_hr_test2}
\end{table}

\begin{table}[t]
    \centering
    \resizebox{0.5\textwidth}{!}{
    \begin{tabular}{ccccccc}
         \toprule
         Method & Presence & Comparison & Rural/Urban & Average Accuracy \\
         \midrule
         RSVQA \cite{lobry2020rsvqa} & 87.46 & 81.50 & 90.00 & 86.32 \\
         EasyToHard \cite{yuan2022easy} & 90.66 & 87.49 & 91.67 &  89.94\\
         Bi-Modal \cite{bazi2022bi} & 91.06 & 91.16 & 92.66 & 91.63 \\
         SHRNet \cite{shrnet} & 91.03 & 90.48 & 94.00 & 91.84 \\
         \midrule
         RSGPT (ours) & \textbf{91.17} & \textbf{91.70} & \textbf{94.00} & \textbf{92.29} \\
         \bottomrule
    \end{tabular}
    }
    \caption{Comparison with State-of-the-arts on the RSVQA-LR Dataset. All Accuracy Scores are Measured by \%.}
    \label{tab_sota_rsvqa_lr_test}
\end{table}

\subsection{Comparison with SOTA}
In this sub-section, we compare RSGPT with state-of-the-art methods across three existing remote sensing image captioning (RSIC) datasets and two existing remote sensing visual question answering (RSVQA) datasets.
\subsubsection{RSIC}
Three different datasets, namely UCM-captions \cite{ucm_sydney_caption}, Sydney-captions \cite{ucm_sydney_caption}, and RSICD \cite{rsicd}, are included in the RSIC comparison experiments.
UCM-captions contains 2,100 high-resolution remote sensing images and 10,500 caption descriptions. Sydney-captions contains 613 images and 3,065 caption descriptions. RSICD contains 10,921 images and 54,605 captions. All three datasets have 5 descriptions for each image.
As the caption descriptions in these datasets are very short and limited in vocabulary, we fine-tuned RSGPT on these datasets to generate output results with similar length and vocabulary as the ground truth captions. 
Specifically, we fine-tuned RSGPT for 5, 5, and 15 epochs on the training sets of UCM-captions, Sydney-captions, and RSICD, respectively, and evaluated on the corresponding test sets.
The evaluation metrics used are typical metrics for image captioning, including BiLingual Evaluation Understudy (BLEU), Recall-Oriented Understudy for Gisting Evaluation (ROUGE\_L), Metric for Evaluation of Translation with Explicit ORdering (METEOR), and Consensus-based Image Description Evaluation (CIDEr). For BLEU, we used n=1,2,3,4 for n-gram precision.
Tables \ref{tab_sota_ucm}, Table \ref{tab_sota_sydney}, and Table \ref{tab_sota_rsicd} show the experimental results.
As shown in the tables, RSGPT significantly outperforms previous methods on most metrics. For example, on BLEU-1, RSGPT surpasses the previous best method by 4.55\%, 3.89\%, and 3.42\% on the UCM-captions, Sydney-captions, and RSICD datasets, respectively.
As can be seen, RSGPT easily surpasses other methods on the existing RSIC datasets with only a few training epochs, demonstrating the powerful ability of large visual-language model.

\subsubsection{RSVQA}
RSVQA-HR \cite{lobry2020rsvqa} and RSVQA-LR \cite{lobry2020rsvqa} datasets are utilized for RSVQA comparison experiments.
RSVQA-HR \cite{lobry2020rsvqa} contains 10,569 high-resolution images and 1,066,316 question-answer pairs, of which 61.5\%, 11.2\%, 20.5\%, and 6.8\% are split into the training set, validation set, test set 1, and test set 2, respectively. There are four question types in this dataset, including presence, comparison, area, and count.
RSVQA-LR \cite{lobry2020rsvqa} contains 772 low-resolution images and 77,232 question-answer pairs, of which 77.8\%, 11.1\%, and 11.1\% are used for training, validation, and testing, respectively. The question types include presence, comparison, rural/urban, and count.
We fine-tuned RSGPT for 5 epochs on the training sets of RSVQA-HR and RSVQA-LR, and evaluated it on the corresponding test sets.
We removed area and count questions during training and testing because the answers to area and count questions are numerical and quantized into multiple categories. For example, in the RSVQA-LR dataset, the answers to counting questions are quantized into the five categories: 0, between 1 and 10, between 11 and 100, between 101 and 1000, and more than 1000.
In addition, we constrained the maximum and minimum lengths of RSGPT's answers to 5 and 2 characters, respectively, because for presence and comparison questions, the ground-truth answers are only "yes" or "no", and for rural/urban questions, the ground-truth answers are only "rural" or "urban".
Table \ref{tab_sota_rsvqa_hr_test1} and Table \ref{tab_sota_rsvqa_hr_test2} show the experimental results on RSVQA-HR test set 1 and test set 2, respectively. The accuracy of each question type and the average accuracy are reported in the tables. On test set 1, RSGPT achieved comparable results to the previous best method SHRNet \cite{shrnet}, while on test set 2, RSGPT outperformed SHRNet on all metrics. Table 3 shows the results on the RSVQA-LR test set, where RSGPT outperformed previous methods on all metrics.
As shown, RSGPT can match or even surpass the SOTA method with only 5 epochs of fine-tuning.
Experiments on 5 datasets of RSIC and RSVQA tasks demonstrated the effectiveness and superiority of RSGPT on multiple vision-language related tasks.

\section{Conclusion}
In this paper, we propose a high-quality dataset, RSICap, to facilitate the development of visual-language models in the field of remote sensing (RS).
RSICap contains 2,585 manually annotated image captions, each of which provides detailed descriptions of the image scene, as well as object information such as color, shape, position, and quantity. It is the most information-rich dataset in the RS field to date, with the longest caption length (an average of 60 vocabularies per caption).
In addition, we also manually annotated an evaluation set, RSIEval, which includes 100 image-caption pairs and 936 visual question-answer pairs, providing a benchmark for the evaluation of visual-language models on the RSIC and RSVQA tasks.
We trained a RS visual-language model, RSGPT, on the RSICap dataset, and evaluated it on RSIEval.
As the model's responses are open-ended, we conducted manual scoring on them. The experimental results show that RSGPT outperforms other models significantly in both tasks.
Furthermore, we evaluated RSGPT across five existing remote sensing datasets, including two RSIC datasets and three RSVQA datasets, and the experimental results demonstrate that RSGPT outperforms the previous state-of-the-art methods in most metrics.

\bibliography{egbib}
\bibliographystyle{ieee}

\end{document}